\documentclass[lettersize,journal]{IEEEtran}
\usepackage{amsmath,amsfonts}
\usepackage{algorithmic}
\usepackage{algorithm}
\usepackage{array}
\usepackage[caption=false,font=normalsize,labelfont=sf,textfont=sf]{subfig}
\usepackage{textcomp}
\usepackage{stfloats}
\usepackage{url}
\usepackage[hidelinks]{hyperref}  
\usepackage{verbatim}
\usepackage{graphicx}
\usepackage{cite}
\begin{document}
\title{Determination of efficiency indicators of the stand for intelligent control of manual operations in industrial production}

\author{Anton Sergeev, Aleksei Soldatov, Victor Minchenkov}

\maketitle

\begin{abstract}
Manual operations remain essential in industrial production because of their flexibility and low implementation cost. However, ensuring their quality and monitoring execution in real time remains a challenge, especially under conditions of high variability and human-induced errors. In this paper, we present an AI-based control system for tracking manual assembly and propose a novel methodology to evaluate its overall efficiency. The developed system includes a multicamera setup and a YOLOv8-based detection module integrated into an experimental stand designed to replicate real production scenarios. The evaluation methodology relies on timestamp-level comparisons between predicted and actual execution stages, using three key metrics: Intersection over Union (IoU), Mean Absolute Scaled Error (MASE), Residual Distribution histograms. These metrics are aggregated into a unified efficiency index ($E_{\text{total}}$) for reproducible system assessment. The proposed approach was validated on a dataset of 120 assemblies performed at different speeds, demonstrating high segmentation accuracy and identifying stage-specific timing deviations. The results confirm the robustness of the control system and the applicability of the evaluation framework to benchmark similar solutions in industrial settings.
\end{abstract}


\section{Introduction}

\IEEEPARstart{T}{he} rapid advancement of machine learning in the 21st century has accelerated the adoption of intelligent technologies in various domains, including industrial production. Although automation and robotics are becoming increasingly prevalent, many complex or safety-critical manual operations remain indispensable. For small and medium-sized manufacturers, full robotic assembly lines often incur prohibitive costs and long payback periods, making manual labor the only viable option. However, human-driven processes are inherently prone to errors, leading to product defects and safety incidents. This requires the development of intelligent monitoring systems capable of regulating manual workflows in real time.

Despite growing interest in this field, most studies focus on isolated components such as detection models or subsystems, neglecting the evaluation of control systems as integrated solutions. The variability of human behavior further complicates the development and assessment of such systems. As a result, there is a lack of standardized methodologies and performance metrics to evaluate the effectiveness of manual operation monitoring. This gap limits both academic progress and practical adoption.

To address these challenges, we present a complete hardware-software system for intelligent control of manual assembly operations and propose a novel methodology for its efficiency evaluation.

The remainder of the paper is organized as follows. Section II reviews related work; Section III defines the problem and research objectives; Section IV describes the experimental setup; Section V details the evaluation methodology; Section VI presents results; and Section VII concludes the study.

\section{Related work}
\IEEEpubidadjcol

\subsection{Control systems}

Automated assembly lines are well-established systems characterized by structured workflows, centralized control, and advanced data processing tools, such as artificial neural networks (ANN), CFD\&POSE descriptors, and boundary object function analysis~\cite{ref1}. These systems integrate robotic manipulators~\cite{ref2,ref3}, computer vision (CV), and comprehensive automation frameworks.

CV technologies are widely used for real-time defect detection and quality inspection in robotic assembly. For example, Yousif et al.\cite{ref4} proposed a digital twin system that identifies components with over 95\% accuracy and corrects errors without halting production. Frustaci et al.\cite{ref5} developed a CV-based inspection system for catalytic converters capable of detecting submillimeter misalignments in 1.3 seconds per part. Another solution~\cite{ref6} integrated RGB-D vision, tactile sensing, and force feedback for autonomous gearbox assembly, achieving a 99.1\% success rate under 225 randomized conditions.

Despite their effectiveness, such robotic systems remain economically inaccessible to many small and medium-sized manufacturers. High integration costs, geopolitical restrictions, and the superior dexterity of human operators in fine-grained tasks limit the adoption of full automation~\cite{ref7}. As a result, manual labor remains prevalent, which requires intelligent monitoring tools for quality assurance.

Recent CV-based systems for manual operation control have shown promise. Bian et al.\cite{ref8} introduced the Smart Connected Worker (SCW) platform that combines object detection, finger tracking, and full body skeleton analysis for additive manufacturing, achieving 99.8\% machine state detection and 92.5\% finger localization. Chen et al.\cite{ref9} proposed a lightweight YOLOv3-based system with CPM pose estimation to monitor repetitive actions, reporting 92.8\% accuracy in action recognition and 82.1\% in repetition counting. Ashourpour et al.~\cite{ref10} applied a YOLOv8-based model to detect defects in a chainsaw assembly line, achieving 99.5\% mAP under challenging conditions.

\subsection{Efficiency evaluation}

Evaluating control systems in robotic environments is relatively straightforward due to the deterministic nature of programmed actions. Standard metrics such as accuracy, precision, recall, and process-level indicators such as cycle time and defect rate are widely used.

However, manual assembly introduces high variability, making traditional evaluation approaches insufficient. Most existing studies assess only individual modules, e.g., object detection or pose estimation, without considering overall system performance. For example, SCW~\cite{ref8} reports high accuracy in component tracking but does not measure its impact on quality control. Similarly, Chen et al.~\cite{ref9} evaluated isolated detection tasks without analyzing the effectiveness at the process level.

The lack of system-level evaluation stems from the nondeterministic nature of human actions: variable timing, motion patterns, and tool use complicate the definition of ground truth and formal metrics. As a result, current approaches are limited to narrow tasks with manually defined workflows.

Notable exceptions include the ViMAT system of Nardon et al.\cite{ref11}, which models the assembly process as a probabilistic state-transition graph, enabling the tracking of complete operation sequences and the detection of deviations from a predefined plan. The approach provides deeper process-level analysis, but depends on manually defined states and rigid task structures, reducing adaptability to variable workflows. A similar method by Oyekan et al.\cite{ref12} employs RGB-D cameras and a state representation inspired by the Hidden Markov Model to infer assembly progress and generate real-time alerts. However, the evaluation focuses on qualitative results and lacks standardized metrics, limiting its applicability to highly variable or loosely structured production environments.

In summary, previous work lacks a unified quantitative methodology for evaluating manual operation control systems as integrated entities. Most approaches either target isolated components or rely on fixed rule-based workflows. This highlights the need for a generalizable evaluation framework, a challenge addressed in this study.

\section{Problem statement}

The development of an AI-based control system for manual operations requires a structured hardware-software platform capable of real-time video processing and performance evaluation. The system must include a vision-based monitoring setup with high-resolution cameras and computing modules to detect operator actions, synchronize them with operational timelines, and identify deviations from expected task sequences. The experimental environment should replicate real production conditions to ensure applicability and robustness.

Unlike automated assembly lines, manual processes involve high variability and are less suited to conventional metrics such as throughput or fault rate. This makes system evaluation more challenging and requires specialized indicators.

To address this, the evaluation methodology must rely on objective and reproducible metrics derived from annotated video data and model predictions. Key performance indicators include Intersection over Union (IoU) for segmentation accuracy, Mean Absolute Scaled Error (MASE) for timing discrepancies, and Residual histograms for assessing temporal prediction balance. Together, these metrics provide a rigorous and interpretable framework for comparing AI-based control systems against traditional manual supervision.

To evaluate the overall effectiveness of the system, relying solely on individual metrics is insufficient. The methodology should incorporate aggregate visual analysis to reveal structural weaknesses and temporal inconsistencies, along with a well-defined strategy to consolidate these insights into a single, interpretable efficiency measure.

\section{AI-based stand for manual operations control}

\subsection{System description}

To enable intelligent monitoring of manual operations, a dedicated experimental stand was developed, replicating realistic industrial conditions. The stand functions as an automated operator’s workstation and is designed for real-time video capture, action recognition, and performance feedback.

The setup includes a horizontal working surface and several fixed Basler acA4024-29uc cameras equipped with C23-0824-5M f8mm lenses. The cameras are positioned 1.3 to 2.5 meters above the table to ensure full coverage of the working area. They capture Full HD video at 30 frames per second and transmit the data stream to a Python-based control application with integrated neural network models.

Video streams are processed in real time by a high-performance computing module running Linux. The hardware configuration includes an NVIDIA 3080 Ti GPU, an Intel Core i5 9th-generation processor, and 16 GB of DDR4 RAM, providing sufficient computational capacity for neural inference and data analysis. The system maintains an average inference rate of 10 fps, ensuring timely detection of operator actions.

\begin{figure}[ht]
\centering
\includegraphics[width=1\linewidth]{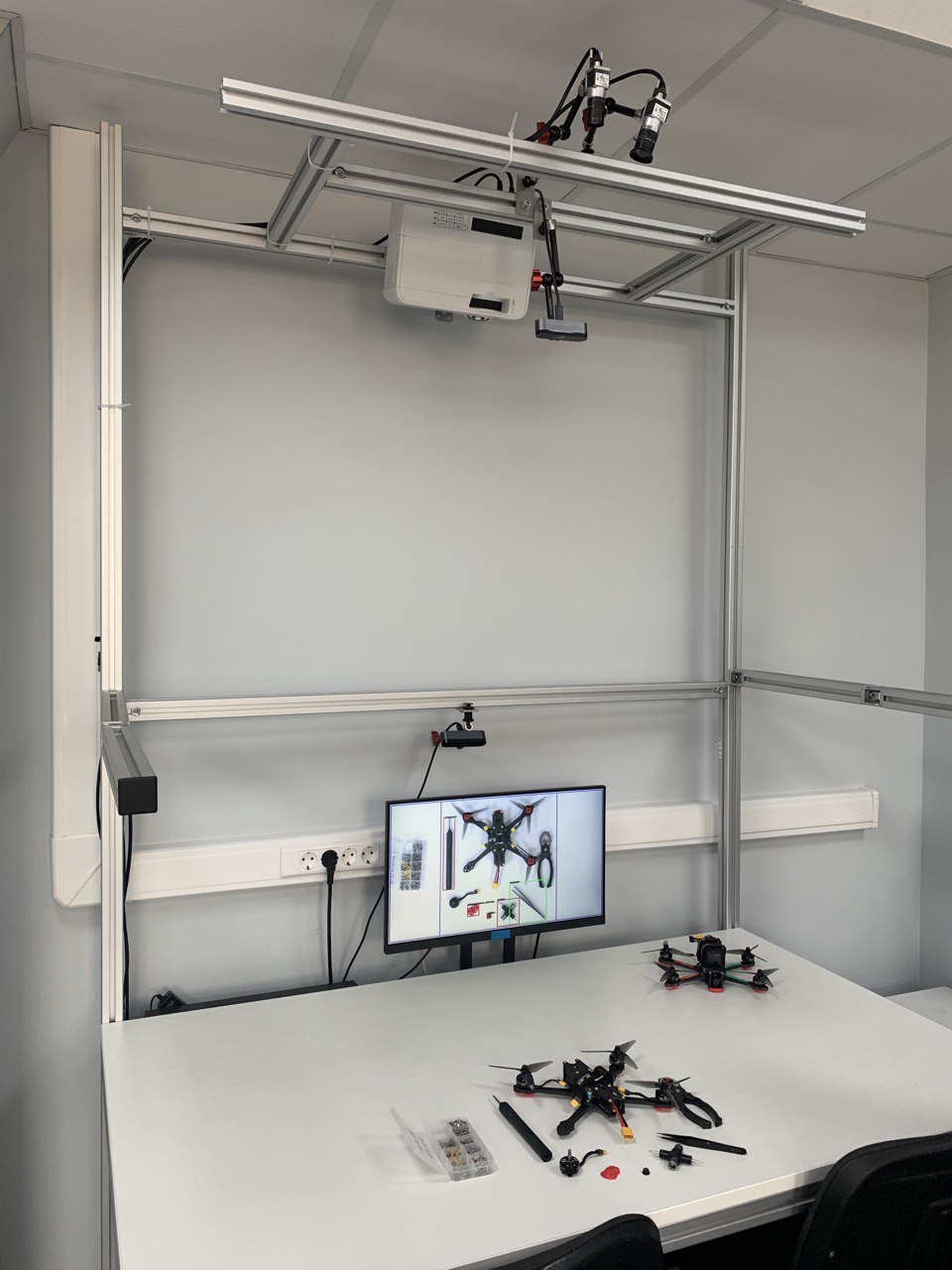}
\caption{The stand for intelligent control and visualization of manual operations}
\label{stand_photo}
\end{figure}

A monitor and projector are integrated into the stand to provide real-time feedback, display system status, and guide the operator during the assembly process. The complete assembled stand is shown in Fig.~\ref{stand_photo}.

\subsection{Assembly process control scenario}

\subsubsection{Algorithm of work}

The assembly process is governed by a strict algorithm that defines the only correct sequence for connecting product components. Each step includes two key actions: placing the required parts in predefined zones and presenting their connection to the cameras.

The system does not proceed to the next step until all conditions for the current stage are satisfied. Informational prompts guide the operator by displaying the required actions on the screen.

After frame processing, the neural network detects the positions of objects and checks their intersection with designated zones. If parts are missing or incorrectly placed, a notification is triggered.

\begin{figure}[ht]
\centering
\includegraphics[width=1\linewidth]{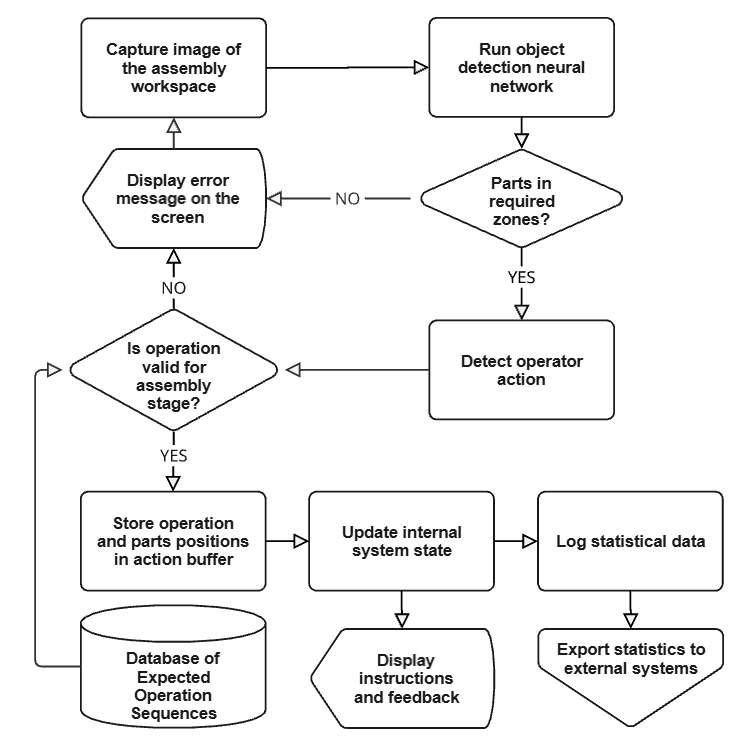}
\caption{System workflow diagram (for any stage)}
\label{workflow}
\end{figure}

To verify proper attachment, the system detects intermediate connections. To reduce computational load, detection is limited to the assembly zone. The model processes the image, returns a list of detected connections with probabilities, and compares them against a confidence threshold. If both primary and auxiliary cameras confirm the correct connection for the current stage, the system proceeds; otherwise, it prompts the operator to repeat the action. The general workflow is illustrated in Fig.~\ref{workflow}.

The system enables the following:
\begin{itemize}
\item real-time error detection during manual operations,
\item reduction of assembly defects and regulatory violations,
\item mitigation of human-factor risks,
\item operator guidance through visual feedback,
\item compliance monitoring and data collection for analysis,
\item improved transparency for decision-making.
\end{itemize}

\subsubsection{Datasets and data collection process}

The YOLOv8 detection model~\cite{ref14,ref15,ref16,ref17} is employed in the proposed control system to identify tools and components during manual operations. Accurate detection requires a diverse and well-annotated training dataset~\cite{ref18,ref19}. To accelerate dataset preparation, we applied automated generation methods~\cite{ref20,ref21,ref22,ref23} based on MediaPipe Hands and OpenCV-based algorithms~\cite{ref24,ref25,ref26,ref27,ref28}, specifically adapted for our task. A detailed description of this approach is provided in~\cite{ref29}. This approach significantly reduced annotation time and improved data diversity, contributing to model reliability in real-world conditions.

The final dataset included 3,000 images across 6 product classes and 2,500 images across 5 tool classes. Augmentation techniques, such as rotation, scaling, color shifts, and Gaussian noise, were applied to improve robustness under varying lighting and environmental conditions.

\subsection{Experiments results: Evaluation of AI models}

The YOLOv8s detection models were trained on the datasets described above and evaluated on separate sets of tools and product components, achieving a mAP@0.5 of 0.98. This result demonstrates the high detection accuracy and robustness of the model in diverse input conditions, supported by effective data augmentation.

Although the detection performance is strong, mAP alone does not reflect the overall efficiency of the control system. To address this, the next section introduces a methodology for system-level evaluation.

\section{Methodology}
\subsection{Dataset preparation}

\begin{figure}[ht]
\centering
\includegraphics[width=1\linewidth]{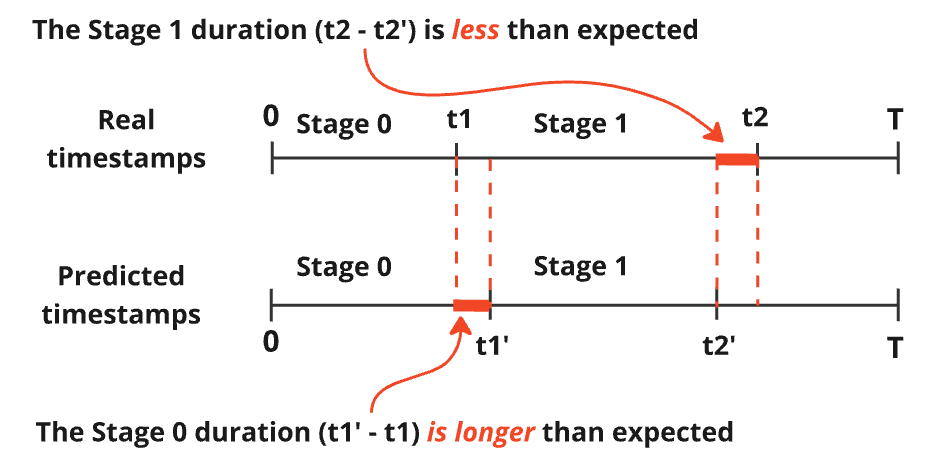}
\caption{Visual comparison of predicted and real timestamps}
\label{iou_1d}
\end{figure}

To enable a standardized and interpretable evaluation of manual operation control systems, we propose a methodology based on timestamp comparison across discrete assembly stages. To conduct the evaluation, we collect a dataset of manually performed assembly sequences. Each sequence is segmented into a fixed number of stages that reflect the production workflow. The ground truth start and end timestamps for each stage are manually annotated from video recordings, while predicted timestamps are extracted from the control system under test (see Fig.~\ref{iou_1d}). 

This approach is applicable to any system where the process can be decomposed into sequential steps — such as part placement, verification, or quality control — regardless of specific implementation details.

To ensure robustness, sequences are grouped by assembly speed. Each group represents a distinct operational condition, enabling performance evaluation at varying execution tempos.

This setup enables a system-level evaluation of temporal accuracy, robustness, and consistency that goes beyond final product quality or isolated model metrics.

\subsection{Evaluation metrics}

The core metrics used to evaluate the system include Intersection over Union (IoU), Mean Absolute Scaled Error (MASE), and Residual histograms. Collectively, these metrics provide a comprehensive assessment of operation segmentation accuracy (via IoU), temporal prediction error (via MASE), and systematic timing deviations (via Residual histograms), enabling a detailed evaluation of control system accuracy, stability, and robustness across all stages of the manual process.

To evaluate the temporal segmentation accuracy of each stage, our methodology applies a one-dimensional (1D) IoU metric, which quantifies the proportion of temporal overlap between predicted and actual intervals. It is computed as follows:

\begin{align}
\label{iou_1d_eq}
\text{IoU} = \frac{\max(0, \min(t2, t2') - \max(t1, t1'))}{\max(t2, t2') - \min(t1, t1')}
\end{align}

This formula covers all overlap scenarios — complete, partial, or none. IoU is computed for each stage individually, and the overall IoU for a given assembly is the average across all stages. The values range from 0 to 1, making it easier to interpret. Low IoU (near 0) suggests failure to detect operation timing boundaries and often implies the need for reconfiguration or exclusion of visually ambiguous stages.

The 1D IoU metric is robust to minor boundary shifts, making it suitable for evaluating variable human actions. It is independent of absolute stage duration and enables consistent comparison between stages, speeds, and different control systems.

To complement IoU, we adopted MASE, a metric originally developed for time series forecasting ~\cite{ref30, ref31}. MASE represents the mean prediction error normalized by typical variation in real durations, making it percentage-based and scale-independent. It is defined as:

\begin{align}
\label{mase_eq}
\text{MASE} = \dfrac{1}{n} \sum_{i=1}^{n} \left| \dfrac{A_i - F_i}{\frac{1}{n-1} \sum_{j=2}^{n} \left| A_j - A_{j-1} \right|} \right|
\end{align}

Here, $A_i$ and $F_i$ are actual and predicted durations. High MASE values indicate undertrained models or weak generalization for specific operations.

To complement individual metric analysis, our methodology includes basic statistical procedures such as correlation analysis and confidence interval estimation. In particular, the Pearson correlation coefficient is used to quantify the relationship between IoU and MASE under different operating conditions. This helps reveal whether poor segmentation accuracy is systematically associated with higher timing errors. In addition, 95\% confidence intervals are computed for both metrics to assess the stability and reliability of system performance between speed groups.

In addition to segmentation accuracy (IoU) and timing prediction error (MASE), we perform residual analysis to examine the distribution and polarity of raw prediction errors at each stage. Unlike aggregated metrics, residuals ($residual = real - predicted$) provide a fine-grained view of whether the system systematically over- or underestimates stage durations. To quantify directional bias, we compute two indicators: \( D \), the proportion of delayed predictions ($residuals < 0$), and \( E \), the proportion of early predictions ($residuals > 0$), both ranging from 0 to 1.

To summarize the balance and concentration of these deviations, we define the \textit{Temporal Balance Score (TBS)}:

\begin{align}
\label{tbs}
\text{TBS} = (1 - |D - E|) \cdot (1 - \max(D, E))
\end{align}

The first factor captures the symmetry between early and late predictions; the second penalizes the dominance of one error type. A higher TBS indicates a more stable and unbiased system, while values near zero suggest strong prediction imbalance.

\subsection{Visualization Procedure}

We visualize key evaluation metrics using stage-wise graphs and histograms to support interpretation and reveal patterns in segmentation accuracy (IoU), timing errors (MASE), and prediction balance (TBS). Stage-wise IoU plots help identify specific operations with weak segmentation performance, while MASE diagrams highlight inconsistencies in temporal prediction across stages and speeds. Residual histograms expose the polarity and spread of prediction errors, allowing detection of systematic bias and temporal instability.

Although the methodology focuses on a compact set of visualizations that are applicable to most scenarios, it remains extensible. Additional plots, such as cumulative timing deviations, operator-specific error trends, or performance heatmaps, can be integrated to support more detailed diagnostics in systems with different architectures or control logic. These visual tools play a critical role in the transformation of raw metric values into actionable insights for system optimization.

\subsection{System Efficiency Index}

We combine all metrics into a single composite score:

\begin{align}
\label{system_efficiency_index}
E_{\text{total}} = \, & \alpha \cdot \overline{\text{IoU}} 
+ \beta \cdot (1 - \overline{\text{MASE}}) \notag \\
& + \gamma \cdot \overline{\text{TBS}}
+ \delta \cdot (1 - \text{CV})
\end{align}

Here, $\text{CV}$ is the coefficient of variation of MASE ~\cite{ref32}. It is used to assess the relative variability of MASE between stages, reflecting the consistency of the timing predictions. A higher CV indicates unstable performance, while lower values suggest a more uniform error distribution. All terms in the formula are normalized to \([0, 1]\). The weights \(\alpha\), \(\beta\), \(\gamma\), and \(\delta\) are selected based on the system priorities and sum to 1. This formulation enables comparisons between systems operating under heterogeneous conditions and allows weight customization to reflect specific production goals.

To interpret metric values, we apply data-driven thresholds derived from the empirical distributions observed in the dataset. For example, acceptable ranges for average IoU, MASE, and CV can be defined using confidence intervals or percentile-based cutoffs. Similarly, thresholds for \( E_{\text{total}} \) can be established to classify the overall efficiency of a system. A final score of \( E_{\text{total}} = 1 \) indicates ideal performance, while \( E_{\text{total}} = 0 \) reflects complete inefficiency.

While the composite score \( E_{\text{total}} \) provides a single interpretable indicator, it can be complemented with individual metric averages (IoU, MASE, TBS, CV) to support more granular comparison between systems, particularly when targeting specific aspects such as segmentation accuracy or temporal stability.

We propose this formula as a generalized framework for evaluating the temporal effectiveness of control systems in manual operations. By adjusting the metric weights and threshold values, this index can be adapted to diverse industrial applications, supporting consistent benchmarking and deployment decisions.

\subsection{Evaluation Workflow}

The evaluation process begins with recording video sequences of manual assemblies performed at different target speeds. Each sequence is manually annotated to define the ground-truth timestamps for individual assembly stages. The system's predicted timestamps are then extracted and compared to the reference data to compute stage-level metrics, including IoU for segmentation accuracy and MASE for timing deviations. Residual values are also calculated to analyze the direction and distribution of prediction errors. Based on these values, additional statistical indicators such as TBS and CV are derived.

To support interpretation and reveal performance patterns, the computed metrics are visualized using aggregated plots that highlight segmentation weaknesses, timing instability, and bias trends across different stages and speed groups. Finally, the overall effectiveness of the control system is quantified using a composite efficiency index ($E_{\text{total}}$), which combines all key metrics into a single interpretable score. This modular and extensible framework supports reproducible evaluation and can be adapted to various system architectures or control strategies in industrial environments.

\section{Results}
The proposed methodology was applied to evaluate our AI-based control system using a dataset of 120 manually performed assemblies. Each assembly sequence consists of 12 sequential stages, and four groups of assembly speeds (30 sequences per group) with target durations of 60, 80, 100, and 120 seconds per assembly (s/assm.) were tested. The actual average durations recorded for each group were approximately 57, 82, 101, and 123 seconds, respectively.

\subsection{Metric-Based Evaluation}

\begin{figure}[ht]
\centering
\includegraphics[width=1\linewidth]{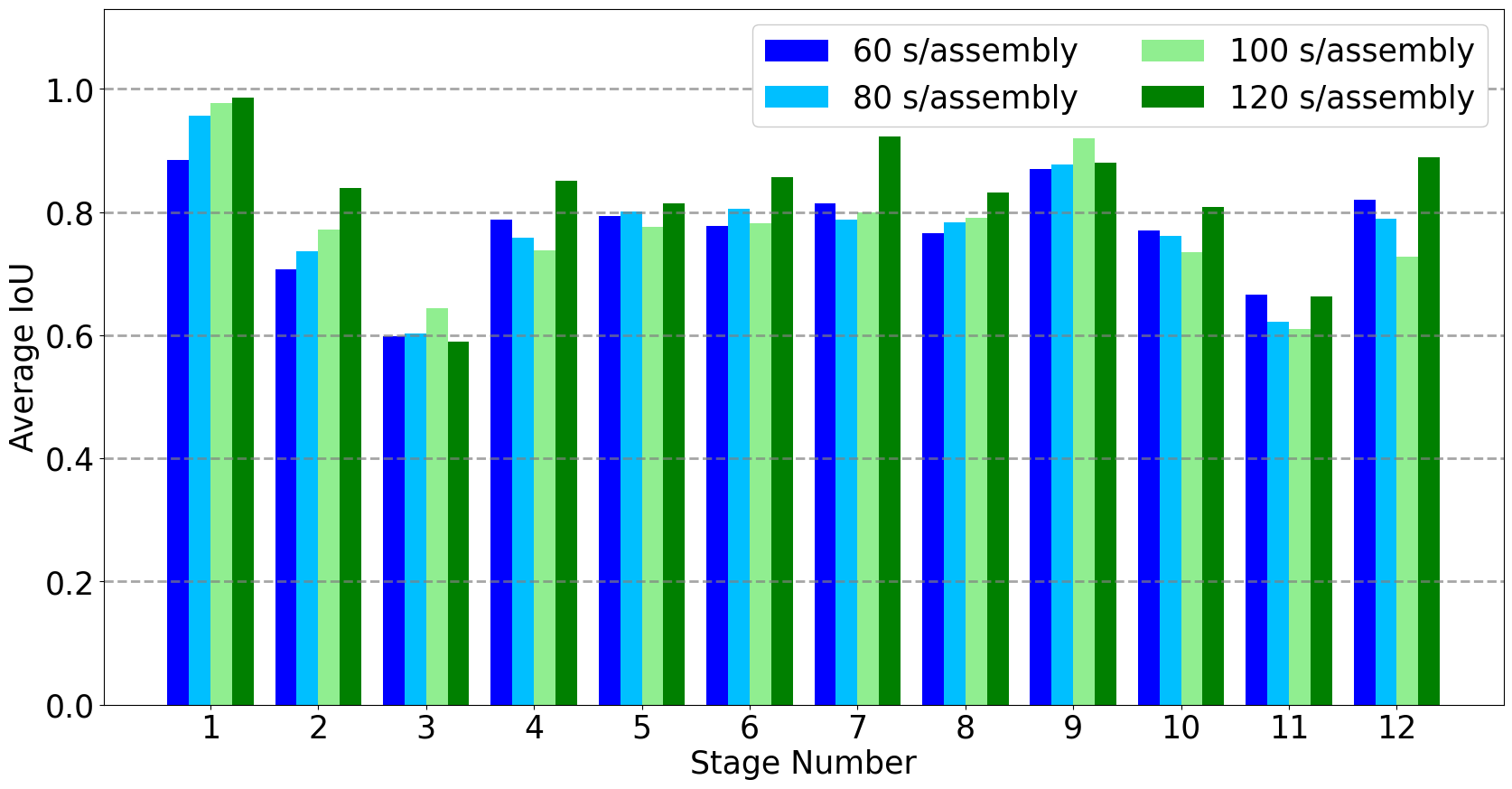}
\caption{Average IoU values for 12 stages and all manufacturing rates}
\label{average_iou}
\end{figure}

The 1D IoU metric was used to assess the alignment between predicted and actual stage durations. As shown in Fig.~\ref{average_iou}, prediction accuracy improves with slower assembly speeds, with the average IoU increasing from 0.77 (60 s/assembly) to 0.84 (120 s/assembly).

\begin{figure}[ht]
\centering
\includegraphics[width=1\linewidth]{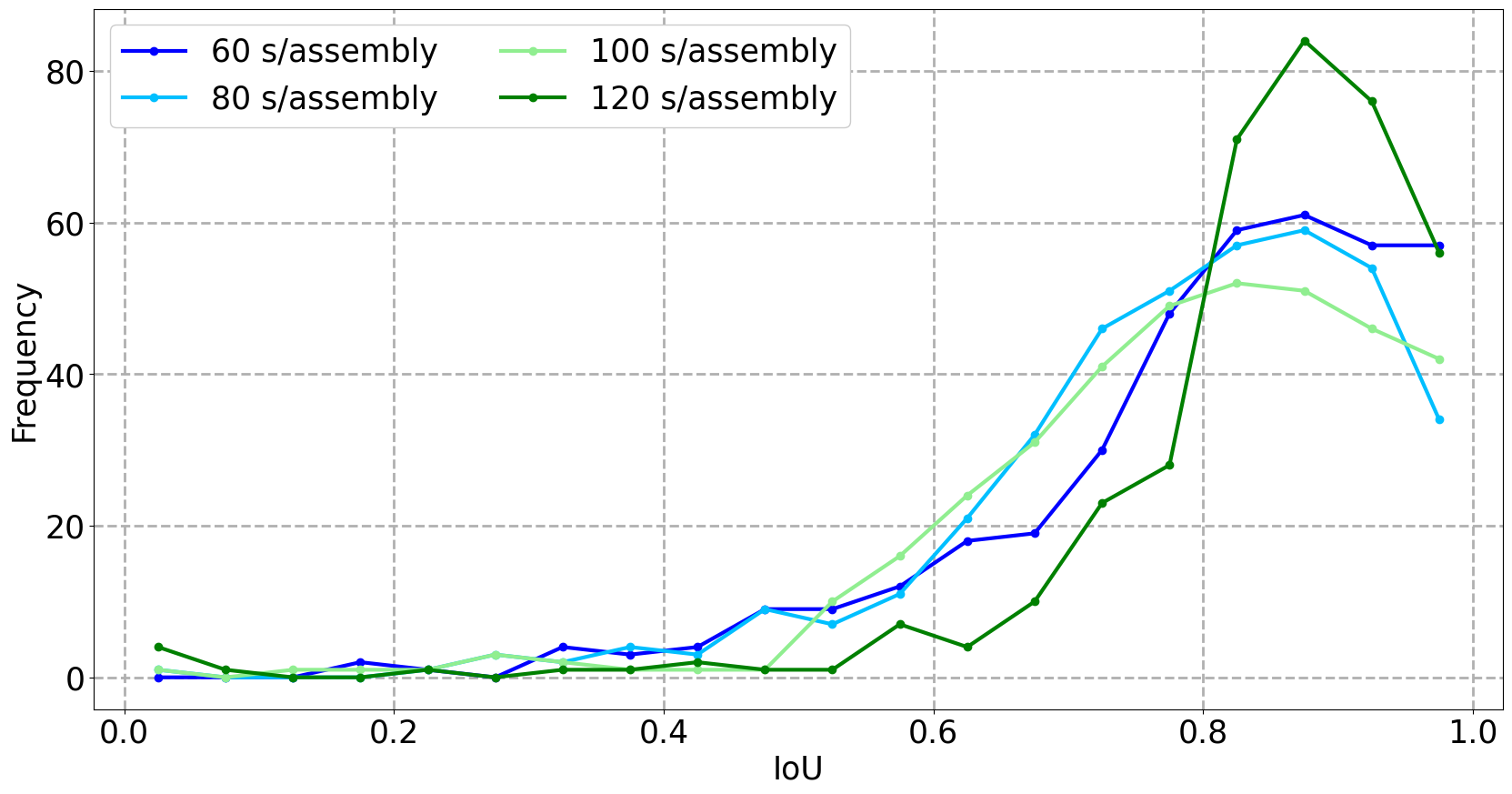}
\caption{Comparison of IoU Distribution for all manufacturing rates}
\label{iou_distribution}
\end{figure}

Fig.~\ref{iou_distribution} confirms this trend: IoU distributions become narrower and shift rightward at lower speeds, indicating improved segmentation stability.

\begin{figure}[ht]
\centering
\includegraphics[width=1\linewidth]{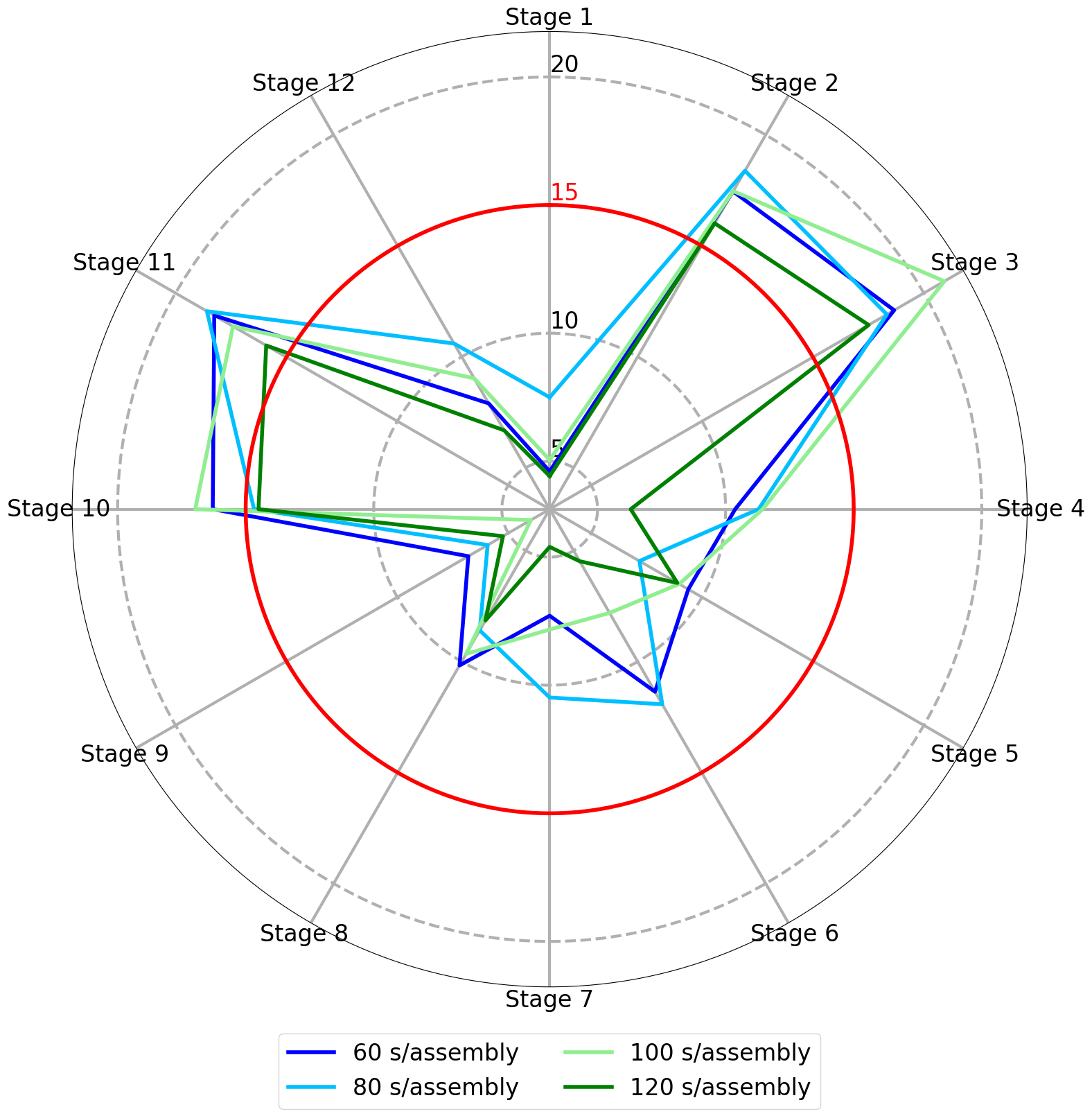}
\caption{MASE values for each stage at different manufacturing rates (in \%)}
\label{mase}
\end{figure}

Stages 3 and 11 consistently show lower IoU values, suggesting difficulties in detecting their boundaries. Fig.~\ref{mase} further reveals elevated MASE values for Stages 2 and 10, indicating inaccurate timing predictions despite acceptable segmentation.

A radar chart highlights timing discrepancies across stages and speeds, with faster assemblies showing higher MASE. On average, MASE decreases from 11.88\% to 8.06\% as speed decreases. A threshold 15\% was defined based on the empirical MASE distribution~\cite{ref33}, corresponding to a standard deviation above the mean (10.67\%, std = 5.3\%).

The correlation analysis confirms a strong negative relationship between IoU and MASE ($r = -0.75$, $p < 0.05$), linking segmentation errors to poor duration prediction. Overall, the system shows reliable performance, but Stages 2, 3, 10, and 11 exhibit localized problems that may require targeted retraining or model adjustments.

\subsection{Residual Error Analysis}

\begin{figure}[ht]
\centering
\includegraphics[width=1\linewidth]{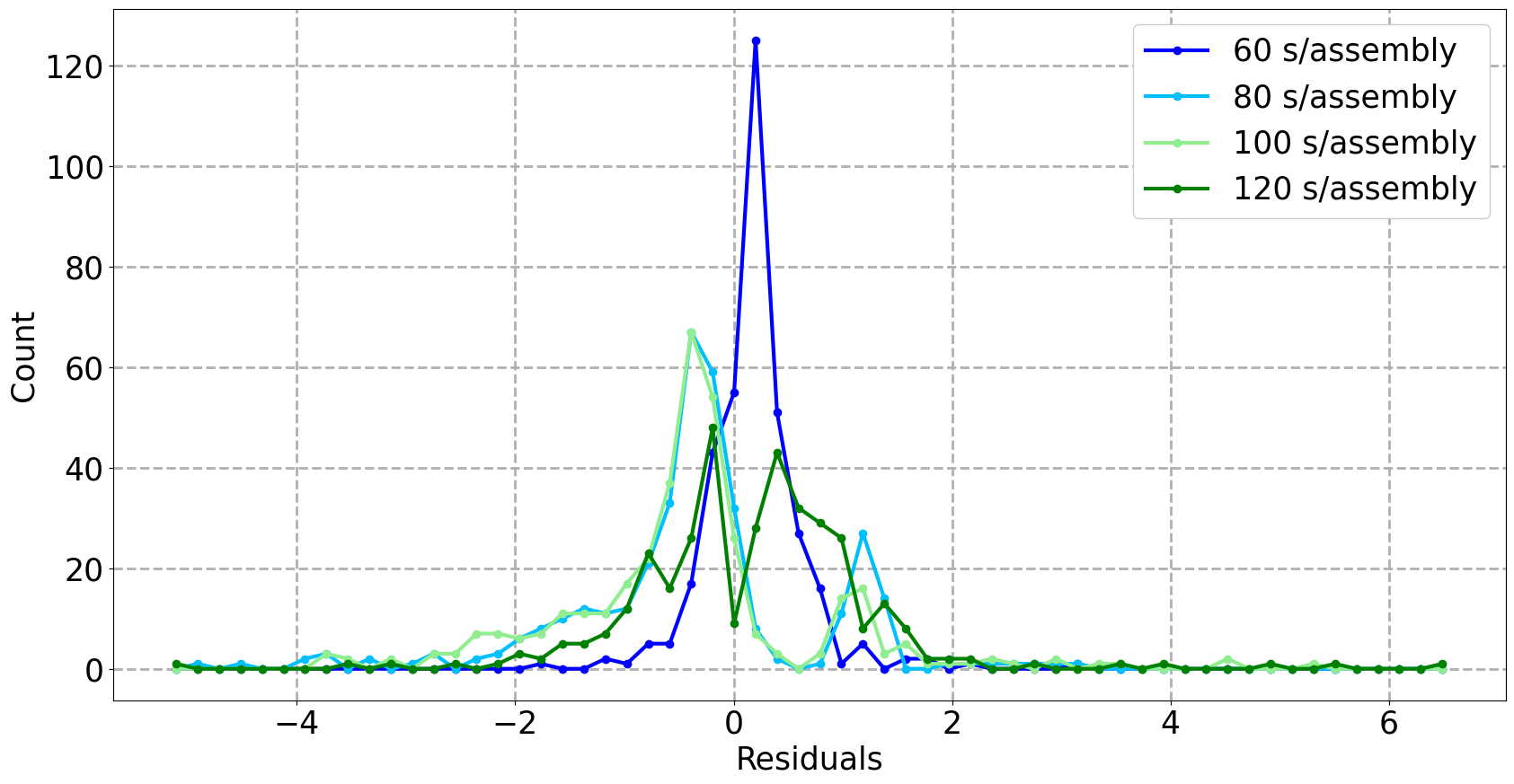}
\caption{Comparison of Residual Distributions for all manufacturing rates}
\label{residuals}
\end{figure}

Residuals, defined as the difference between real and predicted stage durations, were analyzed to detect systematic timing biases. As shown in Fig.~\ref{residuals}, faster speeds produce broader and more skewed distributions, with a tendency toward early predictions ($residuals>0$). At lower speeds, the residuals are more symmetric and concentrated.

Table~\ref{residual_stats} summarizes the residual polarity, skewness, and kurtosis. Mid-speed groups exhibit systematic delays, while high-speed groups tend to anticipate execution. TBS ranges from 0.20 to 0.44, indicating a moderate prediction imbalance between all groups.

\begin{table}[H]
\centering
\caption{Residual Distribution statistics for all manufacturing rates}
\label{residual_stats}
\begin{tabular}{lcccccc}
\textbf{Speed} & \textbf{E} & \textbf{D} & \textbf{TBS} & \textbf{Skewness} & \textbf{Kurtosis} & \textbf{Mode} \\
\hline
60 s/assm.  & 0.42 & 0.10 & 0.39 &  1.06 & 9.13 & 0.08 \\
80 s/assm.  & 0.16 & 0.58 & 0.24 & -0.74 & 3.51 & -0.39 \\
100 s/assm. & 0.15 & 0.62 & 0.20 &  0.71 & 3.76 & -0.32 \\
120 s/assm. & 0.48 & 0.33 & 0.44 &  0.68 & 7.71 & -0.11 \\
\end{tabular}
\end{table}

The distributions confirm two trends: early predictions dominate at high speeds, and delays prevail at intermediate ones. Although residuals are rarely close to zero, their analysis reveals temporal inconsistencies that are not captured by IoU or MASE. To reduce such errors, the model requires calibration per speed mode and retraining on underrepresented patterns.

\subsection{System Efficiency Index Calculation}

To evaluate overall system performance, we computed the composite efficiency index $E_{\text{total}}$ using Eq.~\eqref{system_efficiency_index}, based on average values across all speeds: IoU = 0.79, MASE = 0.11, std(MASE) = 0.05, TBS = 0.32, and $CV = 0.49$. To balance metric contributions, the TBS was scaled as $TBS_{\text{scaled}} = \min(2 \cdot \text{TBS}, 1) = 0.64$.

We evaluated $E_{\text{total}}$ under multiple weight configurations, shown in Table~\ref{E_total_values}. The best performance was observed when higher weights were assigned to IoU and MASE, reflecting the system’s strength in segmentation and average timing accuracy.

\begin{table}[H]
\centering
\caption{System efficiency index with different weight configurations}
\label{E_total_values}
\begin{tabular}{c c c c c c | c}
$\alpha$ & $\beta$ & $\gamma$ & $\delta$ & Target & $E_{\text{total}}$ & Is system efficient? \\
\hline
0.30 & 0.30 & 0.20 & 0.20 & 0.72 & 0.73 & YES \\
0.40 & 0.30 & 0.15 & 0.15 & 0.74 & 0.76 & YES \\
\textbf{0.50} & \textbf{0.30} & \textbf{0.10} & \textbf{0.10} & \textbf{0.76} & \textbf{0.78} & \textbf{YES} \\
0.30 & 0.50 & 0.10 & 0.10 & 0.78 & 0.80 & YES \\
0.20 & 0.20 & 0.40 & 0.20 & 0.68 & 0.69 & YES \\
0.20 & 0.20 & 0.20 & 0.40 & 0.69 & 0.66 & NO \\
0.10 & 0.10 & 0.50 & 0.30 & 0.64 & 0.63 & NO \\
0.60 & 0.20 & 0.10 & 0.10 & 0.75 & 0.77 & YES \\
0.25 & 0.25 & 0.25 & 0.25 & 0.71 & 0.70 & NO \\
\end{tabular}
\end{table}

Increasing the weights assigned to temporal balance and consistency (TBS, CV) led to a lower overall score, reflecting the stage-level variability specific to our system. This effect may differ in alternative configurations or application scenarios, where the relative importance of each metric may vary. These results emphasize the flexibility of the proposed index: by adjusting weights and thresholds, it can be tailored to align with diverse production objectives and quality standards.

\subsection{Results Summary}

The evaluation confirms that the proposed AI-based control system performs reliably in key dimensions. Segmentation accuracy (IoU) improves with slower assembly speeds, reaching an average of 0.84 at 120 s/assm., while temporal accuracy (MASE) remains within acceptable limits, with a mean of 10.67\%. Stages 2, 3, 10, and 11 exhibit localized segmentation or timing issues that may require model refinement.

Residual analysis reveals systematic prediction shifts: early at high speeds and delayed at intermediate speeds. These effects are reflected in residual asymmetry and variability, with TBS values ranging from 0.20 to 0.44 and CV at 49\%. While these deviations are not critical, they highlight areas for targeted calibration.

Overall, the system demonstrates robust performance under variable conditions. The efficiency index $E_{\text{total}}$ consistently exceeds threshold values when prioritizing segmentation and timing metrics, validating the effectiveness of the proposed methodology and confirming its applicability to manual operation monitoring in the real world.

\section{Conclusion}

This work provides a twofold contribution to the field of manual operation monitoring in industrial settings. First, we developed an AI-based control system integrating a multi-camera setup, YOLOv8-based detection, and real-time operator feedback, implemented within a functional stand replicating real production conditions. Second, we proposed a methodology for evaluating such systems based on timestamp-level comparison and key performance metrics, including IoU, MASE, residual analysis, and the derived Temporal Balance Score (TBS). These metrics are combined into a unified efficiency index ($E_{\text{total}}$), allowing a reproducible and interpretable assessment.

Validation in 120 assembly sequences on four speed settings confirmed the system’s robustness, while also highlighting localized segmentation and timing errors. The results demonstrate the applicability of the proposed evaluation framework beyond this specific case, offering a generalizable approach for system-level evaluation of manual operation control.

Future work may focus on improving timing stability, incorporating operator behavior models, and extending the framework to support comparative analysis across different system architectures.

\section*{Acknowledgments}

This work was supported by the grant for research centers in the field of AI provided by the Russian Federation Ministry of Economic Development (agreement 000000C313925P4E0002) and by HSE University (agreement 139-15-2025-009), as well as by the computational resources of the HPC facilities at HSE University.


\begin{thebibliography}{1}
\bibliographystyle{IEEEtran}

\bibitem{ref1}
M. Peña-Cabrera, I. Lopez-Juarez, R. Rios-Cabrera, and J. Corona-Castuera, “Machine vision approach for robotic assembly,” \textit{Assembly Autom.}, vol. 25, no. 3, pp. 204–216, 2005, doi: \nolinkurl{10.1108/01445150510610926}.

\bibitem{ref2}
J.-D. Lee, W.-C. Li, J.-H. Shen, and C.-W. Chuang, “Multi-robotic arms automated production line,” in \textit{Proc. 4th Int. Conf. Control, Autom. Robot. (ICCAR)}, Auckland, New Zealand, 2018, pp. 26–30, doi: \nolinkurl{10.1109/ICCAR.2018.8384639}.

\bibitem{ref3}
KUKA, “Industrial robots from KUKA,” [Online]. Available: \url{https://www.kuka.com/en-us/products/robotics-systems/industrial-robots}

\bibitem{ref4}
I. Yousif, L. Burns, F. El Kalach, and R. Harik, “Leveraging computer vision towards high-efficiency autonomous industrial facilities,” \textit{J. Intell. Manuf.}, vol. 36, no. 5, pp. 2983–3008, 2025, doi: \nolinkurl{10.1007/s10845-024-02396-1}.

\bibitem{ref5}
F. Frustaci, F. Spagnolo, S. Perri, G. Cocorullo, and P. Corsonello, “Robust and high-performance machine vision system for automatic quality inspection in assembly processes,” \textit{Sensors}, vol. 22, no. 8, p. 2839, Apr. 2022, doi: \nolinkurl{10.3390/s22082839}.

\bibitem{ref6}
K. Ota, et al., “Autonomous robotic assembly: From part singulation to precise assembly,” in \textit{Proc. 2024 IEEE/RSJ Int. Conf. Intelligent Robots Syst. (IROS)}, Abu Dhabi, United Arab Emirates, 2024, pp. 13525–13532, doi: \nolinkurl{10.1109/IROS58592.2024.10802423}.

\bibitem{ref7}
Mikol, “Robots vs. human labor in construction: Will automation replace the workforce or create new jobs?,” Oct. 10, 2024. [Online]. Available: \href{https://www.mikolmarmi.com/blogs/news/robots-vs-human-labor-in-construction-will-automation-replace-the-workforce-or-create-new-jobs}{https://www.mikolmarmi.com/blogs/news/robots-vs-human-labor-in-construction-will-automation-replace-the-workforce-or-create-new-jobs}

\bibitem{ref8}
S. Bian, et al., “Machine learning-based real-time monitoring system for smart connected worker to improve energy efficiency,” \textit{J. Manuf. Syst.}, vol. 61, pp. 66–76, 2021, doi: \nolinkurl{10.1016/j.jmsy.2021.08.009}.

\bibitem{ref9}
C. Chen, T. Wang, D. Li, and J. Hong, “Repetitive assembly action recognition based on object detection and pose estimation,” \textit{J. Manuf. Syst.}, vol. 55, pp. 325–333, 2020, doi: \nolinkurl{10.1016/j.jmsy.2020.04.018}.

\bibitem{ref10}
M. Ashourpour, G. Azizpour, and K. Johansen, “Real-time defect and object detection in assembly line: a case for in-line quality inspection,” in \textit{Flexible Automation and Intelligent Manufacturing: Establishing Bridges for More Sustainable Manufacturing Systems (FAIM 2023)}, F. J. G. Silva, A. B. Pereira, and R. D. S. G. Campilho, Eds., Lecture Notes in Mechanical Engineering. Cham, Switzerland: Springer, 2023, pp. 99–106, doi: \nolinkurl{10.1007/978-3-031-38241-3_12}.

\bibitem{ref11}
M. Nardon, et al., “AI-driven visual monitoring of industrial assembly tasks,” arXiv:2506.15285, Jun. 2025. [Online]. Available: \url{https://doi.org/10.48550/arXiv.2506.15285}

\bibitem{ref12}
J. Oyekan, A. Fischer, W. Hutabarat, C. Turner, and A. Tiwari, “Utilising low cost RGB-D cameras to track the real time progress of a manual assembly sequence,” \textit{Assembly Autom.}, vol. 40, no. 6, pp. 925–939, 2020, doi: \nolinkurl{10.1108/AA-06-2018-078}.

\bibitem{ref14}
G. Jocher, “YOLOv5 by Ultralytics, version 7.0,” May 2020. [Online]. Available: \url{https://doi.org/10.5281/zenodo.3908559}

\bibitem{ref15}
M. S. Timoshkin, A. N. Mironov, and A. S. Leontev, “Comparison of YOLOv5 and Faster R-CNN for detecting people in the image in streaming mode,” \textit{Int. Res. J.}, no. 6-1 (120), pp. 137–146, 2022, doi: \nolinkurl{10.23670/IRJ.2022.120.6.020}.

\bibitem{ref16}
J. Nelson, “What is YOLO? The Ultimate Guide [2025],” \textit{Roboflow Blog}, Jan. 9, 2025. [Online]. Available: \url{https://blog.roboflow.com/guide-to-yolo-models/}

\bibitem{ref17}
J. Solawetz and Francesco, “What is YOLOv8? A Complete Guide,” \textit{Roboflow Blog}, Oct. 23, 2024. [Online]. Available: \url{https://blog.roboflow.com/what-is-yolov8/}

\bibitem{ref18}
J. Tremblay, et al., “Training deep networks with synthetic data: Bridging the reality gap by domain randomization,” in \textit{Proc. IEEE Conf. Comput. Vis. Pattern Recognit. Workshops (CVPRW)}, 2018, pp. 969–977.

\bibitem{ref19}
A. Prakash, et al., “Structured domain randomization: Bridging the reality gap by context-aware synthetic data,” in \textit{Proc. 2019 Int. Conf. Robot. Autom. (ICRA)}, Montreal, QC, Canada, 2019, pp. 7249–7255, doi: \nolinkurl{10.1109/ICRA.2019.8794443}.

\bibitem{ref20}
L. Block, A. Raiser, L. Schön, F. Braun, and O. Riedel, “Image-Bot: Generating synthetic object detection datasets for small and medium-sized manufacturing companies,” \textit{Procedia CIRP}, vol. 107, pp. 434–439, 2022, doi: \nolinkurl{10.1016/j.procir.2022.05.004}.

\bibitem{ref21}
Y. Ishida and H. Tamukoh, “Semi-automatic dataset generation for object detection and recognition and its evaluation on domestic service robots,” \textit{J. Robot. Mechatron.}, vol. 32, no. 1, pp. 245–253, Feb. 2020, doi: \nolinkurl{10.20965/jrm.2020.p0245}.

\bibitem{ref22}
H. J. Jo, D. Kim, and J. B. Song, “Automatic dataset generation of object detection and instance segmentation using Mask R-CNN,” \textit{J. Korea Robot. Soc.}, vol. 14, no. 1, pp. 31–39, 2019, doi: \nolinkurl{10.7746/jkros.2019.14.1.031}.

\bibitem{ref23}
X. Lin, L. Xu, and Q. Wang, “Automatic dataset generation for specific object detection,” in \textit{Proc. 2022 IEEE Int. Conf. Image Process. (ICIP)}, Bordeaux, France, 2022, pp. 3076–3080, doi: \nolinkurl{10.1109/ICIP46576.2022.9897741}.

\bibitem{ref24}
OpenCV.org, “OpenCV: OpenCV-Python Tutorials,” OpenCV Documentation, Jul. 7, 2025. [Online]. Available: \url{https://docs.opencv.org/4.x/d6/d00/tutorial_py_root.html}

\bibitem{ref25}
Google, “Hand landmarks detection guide | Google AI Edge | Google AI for Developers,” ai.google.dev, 2025. [Online]. Available: \url{https://ai.google.dev/edge/mediapipe/solutions/vision/hand_landmarker}

\bibitem{ref26}
D. D. Li and X. Y. Liu, “Research on MVP design pattern modeling based on MDA,” \textit{Procedia Comput. Sci.}, vol. 166, pp. 51–56, 2020, doi: \nolinkurl{10.1016/j.procs.2020.02.012}.

\bibitem{ref27}
L. Chen and X. Yang, “Research of applying MVC pattern in distributed environment,” \textit{Jisuanji Gongcheng / Comput. Eng.}, vol. 32, no. 19, pp. 62–64, 2006.

\bibitem{ref28}
F. Zhang, et al., “Mediapipe hands: On-device real-time hand tracking,” arXiv:2006.10214, Jun. 2020. [Online]. Available: \url{https://doi.org/10.48550/arXiv.2006.10214}

\bibitem{ref29}
A. Sergeev, V. Minchenkov, A. Soldatov, S. Lukashov, and Y. Mazikov, “Method of automatic images datasets sampling for the manual operations control systems,” in \textit{Proc. 2023 XVIII Int. Symp. Problems Redundancy Inf. Control Syst. (REDUNDANCY)}, Moscow, Russia, 2023, pp. 194–199, doi: \nolinkurl{10.1109/Redundancy59964.2023.10330200}.

\bibitem{ref30}
A. Ahuja, “Mean Absolute Scaled Error (MASE) in forecasting,” Medium, Jan. 11, 2021. [Online]. Available: \url{https://medium.com/@ashishdce/mean-absolute-scaled-error-mase-in-forecasting-8f3aecc21968}

\bibitem{ref31}
SAP, “Error Measures,” SAP Help Portal: SAP Integrated Business Planning. [Online]. Available: \url{https://help.sap.com/docs/SAP_INTEGRATED_BUSINESS_PLANNING/feae3cea3cc549aaa9d9de7d363a83e6/32cc1c552770a20ce10000000a44176d.html?locale=en-US}

\bibitem{ref32}
A. Hayes, “Coefficient of Variation: Definition and How to Use It,” Investopedia, May 22, 2025. [Online]. Available: \url{https://www.investopedia.com/terms/c/coefficientofvariation.asp}

\bibitem{ref33}
Y. Salimi, D. Domingo-Fernández, M. Hofmann-Apitius, and C. Birkenbihl, “Data-driven thresholding statistically biases ATN profiling across cohort datasets,” \textit{The Journal of Prevention of Alzheimer’s Disease}, vol. 11, no. 1, pp. 185–195, 2024, doi: \nolinkurl{10.14283/jpad.2023.100}.

\bibitem{ref34}
S. Nevil, “Z-Score: Meaning and Formula,” Investopedia, Mar. 11, 2025. [Online]. Available: \url{https://www.investopedia.com/terms/z/zscore.asp}

\bibitem{ref35}
P. Lou, J. Li, Y. Zeng, B. Chen, and X. Zhang, “Real-time monitoring for manual operations with machine vision in smart manufacturing,” \textit{Journal of Manufacturing Systems}, vol. 65, pp. 709–719, 2022, doi: \nolinkurl{10.1016/j.jmsy.2022.10.015}.

\bibitem{ref36}
A. V. Sergeev, V. O. Minchenkov, A. V. Soldatov, Y. A. Mazikov, and V. V. Kakurin, “Outliers resistant image classification by anomaly detection,” \textit{Advances in Artificial Intelligence and Machine Learning}, vol. 5, no. 1, p. 191, Mar. 2025, doi: \nolinkurl{10.54364/AAIML.2025.51191}.

\end{thebibliography}
\end{document}